\documentclass[10pt]{article} 
\usepackage[preprint]{tmlr}

\usepackage{amsmath,amsfonts,bm}

\def\eqref#1{equation~\ref{#1}}

\def\1{\bm{1}}

\DeclareMathAlphabet{\mathsfit}{\encodingdefault}{\sfdefault}{m}{sl}
\SetMathAlphabet{\mathsfit}{bold}{\encodingdefault}{\sfdefault}{bx}{n}

\newcommand{\sigmoid}{\sigma}

\usepackage{hyperref}
\usepackage{url}
\usepackage{graphicx}
\usepackage{subcaption}  
\usepackage{booktabs}
\usepackage{caption} 

\title{Jet: A Modern Transformer-Based Normalizing Flow}

\author{\name Alexander Kolesnikov\thanks{Equal contribution.} \email akolesnikov@google.com \\
      Google DeepMind
      \AND
      \name Andr\'e Susano Pinto\footnotemark[1] \email andresp@google.com \\
      Google DeepMind
      \AND
      \name Michael Tschannen\footnotemark[1] \email tschannen@google.com\\
      Google DeepMind
      }

\def\month{MM}  
\def\year{YYYY} 
\def\openreview{\url{https://openreview.net/forum?id=XXXX}} 

\begin{document}

\maketitle

\begin{abstract}
In the past, normalizing generative flows have emerged as a promising class of generative models for natural images.
This type of model has many modeling advantages: the ability to efficiently compute log-likelihood of the input data, fast generation and simple overall structure.
Normalizing flows remained a topic of active research but later fell out of favor, as visual quality of the samples was not competitive with other model classes, such as GANs, VQ-VAE-based approaches or diffusion models.
In this paper we revisit the design of the coupling-based normalizing flow models by carefully ablating prior design choices and using computational blocks based on the Vision Transformer architecture, not convolutional neural networks.
As a result, we achieve state-of-the-art quantitative and qualitative performance with a much simpler architecture.
While the overall visual quality is still behind the current state-of-the-art models, we argue that strong normalizing flow models can help advancing research frontier by serving as building components of more powerful generative models.
\end{abstract}

\section{Introduction}

In this paper we explicitly do not attempt to devise the new state-of-the art image modeling approach or propose a new paradigm. Instead, we revisit the long known but recently neglected class of models for generative modeling: coupling-based normalizing flows.
Normalizing flows have important capabilities that make them a useful tool for modern generative modeling.

On a high-level, a normalizing flow model learns a bijective (and thus invertible) mapping $g$ from the input space to the latent space, where the latent space follows a simple distribution, e.g. Gaussian distribution.
A complex bijective transformation $g$ can be constructed by stacking multiple \textit{coupling blocks}, which are bijective and invertible in closed form by design and are parametrized by deep neural networks.

Normalizing flow models can be directly trained by computing data log-likelihood in the simple (e.g. Gaussian) latent space after the learnable and differentiable mapping $g$ is applied to transform training examples.
For data generation, the inverse transformation $g^{-1}$ is readily available, which can be used to map easy-to-sample Gaussian latent space to the samples from the target distribution.
The two explicit, differentiable and losseless mappings $g$ and $g^{-1}$ can be used as building blocks for more complex generative systems.
For example, normalizing flows are used as a critical component of more complex systems: recent examples include~\citep{chen2016variational,kingma2016improved,tschannen2023givt} and~\citep{tschannen2024jetformer} that leverage normalizing flows to facilitate image modeling with VAEs and autoregressive transformers, respectively.
This motivates us to revisit the normalizing flow model class.

In the mid-to-late 2010s normalizing flow models were a topic of active research.
NICE~\citep{dinh2014nice} was the early normalizing flow model for images.
It introduced the main building block behind normalizing flow models: an additive coupling layer.
RealNVP~\citep{dinh2017realnvp} improved over NICE by introducing multiscale architecture and affine coupling layers that additionally perform a scaling transformation.
Subsequently, in Glow~\citep{kingma2018glow}, authors introduce two more components: an invertible dense layer and specialized  activation normalization layer.
Finally, Flow++~\citep{ho2019flow++} shows improvements from using dequantization flow trick and generalized variant of the affine coupling block.

In this paper we mainly concentrate on revisiting optimal design for the normalizing flow models. We focus on both performance and simplicity of the final model. We built on top of the prior literature and put all components to a careful scrutiny. Our final model has a radically simpler design and only relies on the plain affine coupling blocks parametrized by the Vision Transformer model. Our key contributions can be summarized as follows:
\begin{itemize}
    \item We use Vision Transformer building blocks instead of convolutional neural networks, which leads to a significant performance improvement.
    \item We drastically simplify the overall architecture by eliminating many components from prior models:
    \begin{itemize}
        \item No multiscale components and early factored-out channels
        \item No invertible dense layers
        \item No ``activation normalization'' layers
        \item No dequantization flow
        \item No generalized coupling transformation
    \end{itemize}
    \item At the same time, we achieve SOTA results in terms of negative log-likelihood across the board on common image benchmarks.
    \item Additionally we demonstrate that transfer learning can be used to tame overfitting.
\end{itemize}

\section{Method}

\begin{figure}[t]
    \centering
    \includegraphics[width=\textwidth]{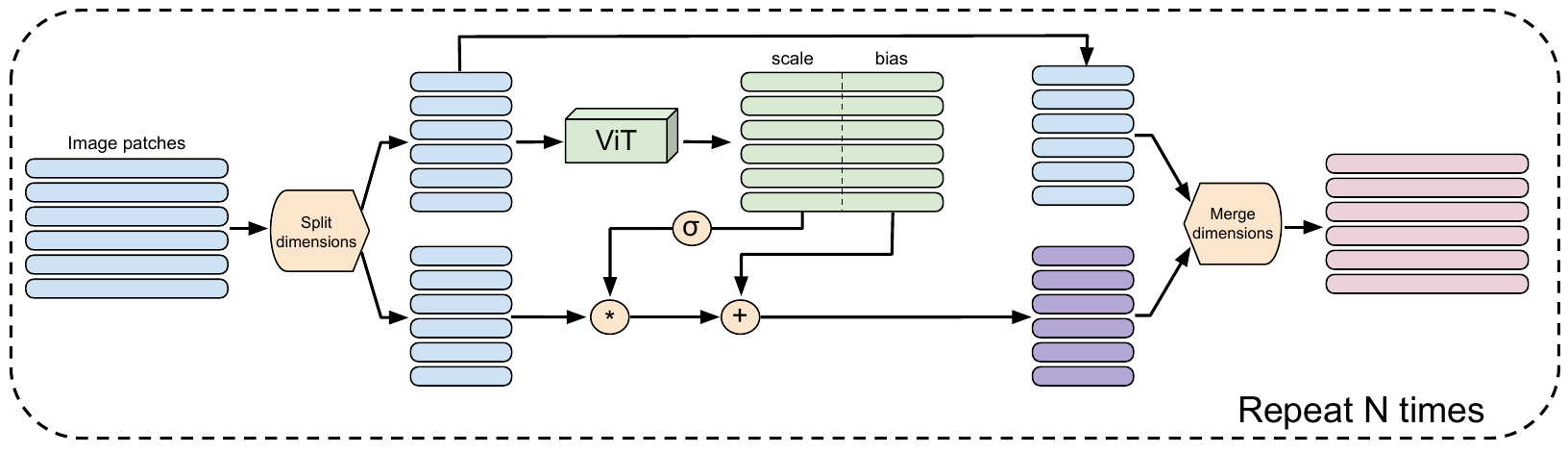}
    \caption{Overview of the Jet model. The dashed box contains a coupling layer computing an affine transform from one half of the input dimensions (patches or features) and then applying it to the other half of the input dinensions. The full model is obtained by stacking $N$ such invertible coupling layers.}
    \label{fig:jet-scheme}
\end{figure}

In the section we introduce the Jet model. We first introduce the architecture, then describe training procedure and important implementation details.

\subsection{Jet model}

\begin{figure}[t]
\centering
\begin{subfigure}[t]{0.48\columnwidth}
\includegraphics[width=1\columnwidth]{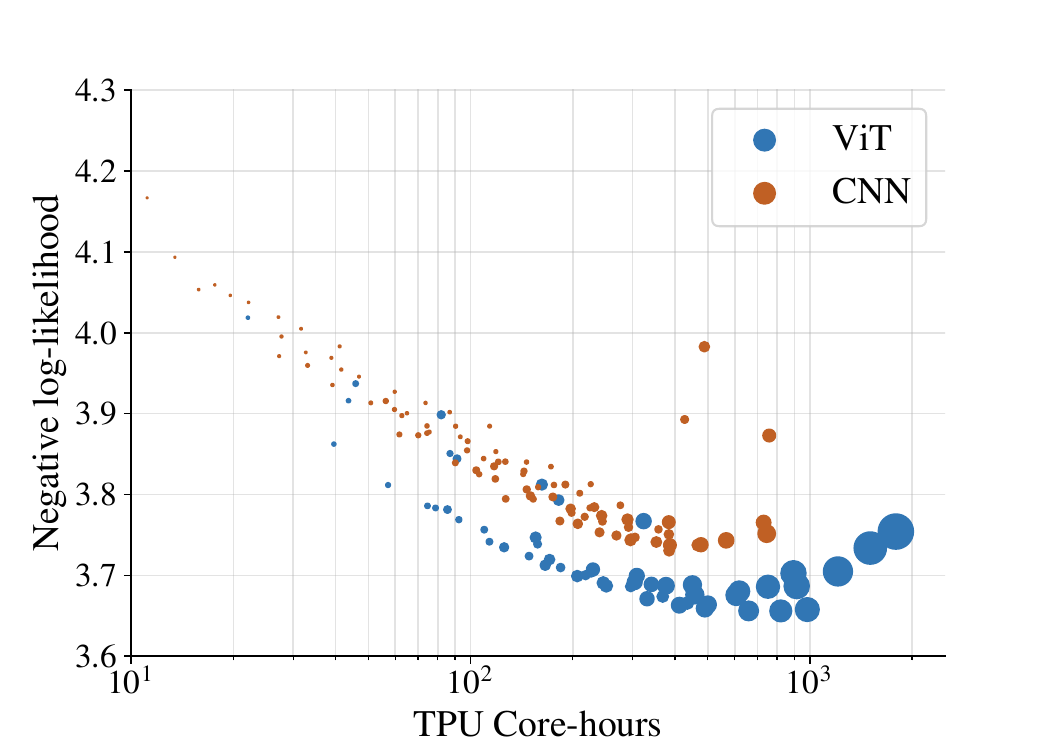}\caption{}
\label{fig:ablate-vit-cnn}
\end{subfigure}
\hspace{0.01\columnwidth}
\begin{subfigure}[t]{0.48\columnwidth}
\includegraphics[width=1\columnwidth]{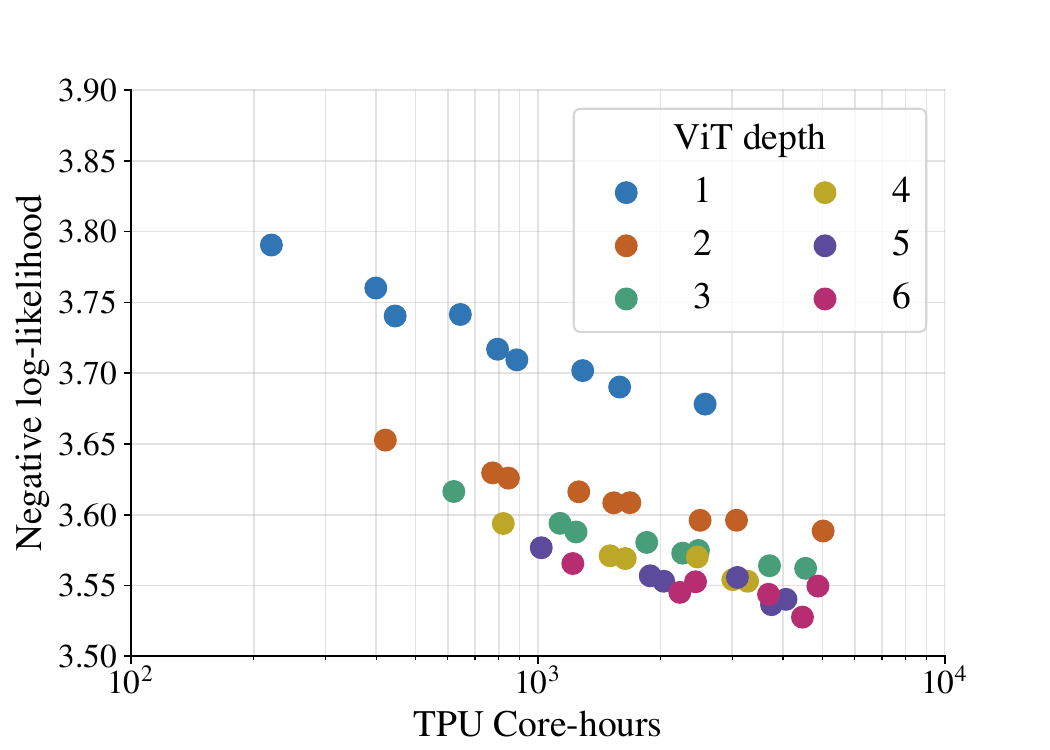}\caption{}
\label{fig:block-depth}
\end{subfigure}
\caption{
Effect of different architecture design choices on the validation NLL (in bits per dimension), as a function of training compute.
Figure~\ref{fig:ablate-vit-cnn}: Results on ImageNet-1k $64\times64$ for CNN vs ViT blocks (the marker size is proportional to the model parameter count). ViT blocks clearly outperform CNN blocks for a given training compute budget.
Figure~\ref{fig:block-depth}: Results on ImageNet-21k $32\times32$ for different ViT depths. Increasing the block depth leads to improved results up to depth 5.
}
\label{fig:ablations}
\end{figure}

The Jet model has a very simple high-level structure. First, the input image is split into $K$ flat patches, flattened into a sequence of vectors, and then we repeatedly apply affine coupling layers~\citep{dinh2017realnvp}. We illustrate the Jet model architecture in Figure~\ref{fig:jet-scheme}.

The input to a coupling layer is $x \in \mathbb{R}^{K \times 2d}$, where $K$ is the total number of patches and $2d$ is the number of dimensions within each patch after flattening›.
We then apply a dimension splitting transformation, producing $x_1, x_2 \in \mathbb{R}^{K \times d}$.
The first part, $x_1$, is not modified, yielding $y_1$.
The second part, $x_2$ is modified with element-wise affine transformation $y_2 = (x_1 + b) \cdot s$, where $s, b \in \mathbb{R}^{K \times d}$.
Scaling factor $s$ and bias $b$ are functions of $x_1$.
We use a single Vision Transformer (ViT) backbone, $f: \mathbb{R}^{K \times d} \rightarrow \mathbb{R}^{K \times 2d}$, which maps $x_1$ to scale and bias, effectively implementing the functions $s(x_1)$ and $b(x_1)$.
Finally, we merge $y_1$ and $y_2$ by applying the inverse of the splitting transformation that was applied to $x$, and arrive at the output $y \in \mathbb{R}^{K \times 2d}$.

To stabilize training we additionally apply sigmoid transformation, $\sigma$, that squashes the scale coefficient to $(0, 1)$ range and then multiply it by a fixed constant $m$ to extend the range to $(0, m)$. In practice we always set $m=2$, which allows to amplify or suppress the intermediate activations, while keeping activations in a reasonable range when stacking many coupling layers. 

Overall, a single coupling block can be formalized as follows:
\begin{align*}
    &y_1 = x_1 \\
    &y_2 = (x_2 + b(x_1)) \cdot \sigmoid(s(x_1)) \cdot m
\end{align*}

As we show below in Section~\ref{sec:inverse}, this transformation can be inverted for arbitrary functions $b(x_1)$ and $s(x_1)$.

\subsection{Training and optimization objective.}

We now recall the fundamentals of the normalizing flow model training. The key assumption behind the normalizing flows is that the target density $p(x)$ can be modeled as a simple distribution, $p_{\mathcal{N}}(z)$ (e.g. standard Gaussian) after using a bijective transformation $g\colon x \rightarrow z$ to transform the original data. Applying the ``change of variable'' identity from the basic probability calculus we obtain a tractable model for the  probability density function:
\begin{align*}
    &p(x) = p_{\mathcal{N}}(g(x)) \left| \mathrm{det} \dfrac{\partial g(x)}{\partial x^\top} \right|.
\end{align*}

We implement function $g$ with the Jet model described above. The determinant of the derivative, $\dfrac{\partial g(x)}{\partial x^\top}$, is equal to the product of determinant of the derivatives of individual coupling layers. And for each coupling block the determinant computation is trivial and equal to the product of all scaling factors in the affine transform. See~\citep{dinh2017realnvp} for details on the derivation. 
In practice we maximize data log-likelihood, so the optimization objective is
\begin{align*}
    -\frac12 \sum\limits_{i}^{} \left[g(x)_i^2 + \log(2 \pi)\right] + \sum\limits_{\ell}^{} \sum\limits_{i}^{} \left[ \log \sigmoid(s_{i}^{\ell})  + \log m \right ],
\end{align*}
where we use index $i$ to iterate over all dimensions of an array inside the sum, and upper index $\ell$ to iterate over total number of blocks. The first sum term arises from taking the logarithm of the standard Gaussian density, applied to every output dimension. The second sum term is the log determinant, which is computed as sum of logarithms of all scale values across all layers, plus logarithm of the fixed $m$ multiplier.

In practice we normalize the above objective by the total number of dimensions and we keep all constant terms. Even though it does not affect optimization, by doing this we obtain a log-likelihood estimate, which can be interpreted as ``bit-per-dimension'' (bpd) (assuming the base of the logarithm is 2). For example, if the objective value is $3.1$, it means that the average uncertainty per input dimension is $3.1$ bits. A uniformly random image model should yield approximately $8$ bpd. This is useful when comparing models across different modeling classes and for various correctness checks.

Note that to correctly model discrete densities (e.g. images where pixels are discrete-valued) we employ a standard dequantization procedure described in~\citep{theis2015note}. In practice, it means that we add a random $[0, 1]$ uniform noise to input images, which consist of discrete values in the range $[0, 255]$. We perform this dequantization procedure both during training and evaluation.

\subsection{Inverse transformation and image generation}
\label{sec:inverse}
It is easy to obtain the inverse of the above transformation in closed form. Notably, the ViT function that computes bias and scale terms does not need to be inverted:
\begin{align*}
    &x_1 = y_1 \\
    &x_2 = \frac{y_2}{\sigmoid(s(x_1)) \cdot m} - b(x_1)
\end{align*}

The computational complexity of computing the inverse is exactly the same as computing the normalizing flow itself. New images can be sampled by first sampling the target density (i.e. Gaussian noise) and then applying the inverse transformation.

\subsection{Initialization}

Careful initialization is essential for training a deep normalizing flow model with a large number of coupling blocks. We employ a simple yet very effective initialization scheme.
The final linear projection of the ViT $f$ is initialized with zero weights. As a result, predicted bias values, $b(x_1)$ are 0. The scale values are equal to $\sigmoid(0) = 0.5$. When we set $m=2$, then the all scaling factors become equal to $1$.

As a result, the Jet model behaves as identity function at initialization. Empirically, we find this sufficient to ensure stable optimization in the beginning of training. Consequently, we do not need to add ``ActNorm'' layer that is commonly used in the normalizing flow literature to achieve a similar effect.

\subsection{Dimension splitting}
\label{sec:dimension-split}

We explore various options for channel spitting. One option is to perform channel-wise split, by splitting the channels of each image patch into the two equal groups~\citep{dinh2014nice, kingma2018glow}. The splitting is random within each coupling layer and is fixed ahead of time (independently for each layer). This is a simple-to implement-strategy that ensures diverse channel mixing.

We also implement various splitting strategies to facilitate spatial mixing. To this end, we explore 3 strategies that respect images' $2$D topology: row-wise alternating patch splitting, column-wise alternating patch-splitting, and the ``checkerboard'' splitting~\citep{dinh2017realnvp}. See Section~\ref{sec:exp} for the empirical investigation of performance for various design choices of splitting operations.

\textbf{Splitting implementation details.} The natural way to perform channel splitting is to use array indexing operations. However, indexing can be slower than matrix multiplication on modern accelerators. Thus, we implement channel splitting as matrix multiplication with precomputed $0$ and $1$ (frozen) weight matrices. A nice by-product of this approach is that dimension merging (inverse of splitting) can be trivially implemented as matrix multiplications with the transposed weight matrices. 

\textbf{Numerical precision considerations.} Matrix multiplications on modern accelerators often implicitly uses half-precision for multiplying individual values, while accumulating the result using full \texttt{float32} precision. We note that such loss in precision may lead to numerical issues, as, for example, the uniform $[0, 1]$ dequantization noise loses most of its entropy and leads to overly optimistic log-likelihood estimates. To avoid this, we enable full precision matrix multiplication mode when splitting/merging the dimensions. Note that we still use fast default settings when computing the ViT function $f$ inside coupling layers. So the overall efficiency drop from enabling full precision for splitting/merging is negligible overall.

\section{Experiments}\label{sec:exp}

Throughout the paper, we keep our experimental setup simple and unified. For the ViT architecture, we follow the original paper~\cite{dosovitskiy2020image}. Our ViT models inside the coupling layers do not have initial patchification or final pooling and linear projections. Unless stated otherwise, we set the patch size such that the total number of patches is equal to $256$.

For the optimizer we use AdamW~\citep{loshchilov2017fixing}. We set second momentum $\beta_2$ parameter to 0.95 to stabilize training.  We use a cosine learning rate decay schedule.

\textbf{Datasets.} We perform experiments on three datasets: Imagenet-1k, Imagenet-21k and CIFAR-10, across two input resolutions: $32\times32$ and $64\times64$ (except for CIFAR-10). To downsample Imagenet-1k images we follow the standard protocol~\citep{chrabaszcz2017downsampled} to ensure a correct comparison to the prior art (i.e. we use the preprocessed data provided by~\citep{chrabaszcz2017downsampled} where available). To downsample Imagenet-21k images we use \texttt{TensorFlow} resize operation with method set to \texttt{AREA}\footnote{\url{https://www.tensorflow.org/api_docs/python/tf/image/resize}}. For CIFAR-10 we use the original dataset resolution. Importantly, to make sure our results are comparable to the literature, we do not perform any data augmentations.

\subsection{Main results}

\begin{figure}[t]
    \centering
    \includegraphics[width=\textwidth]{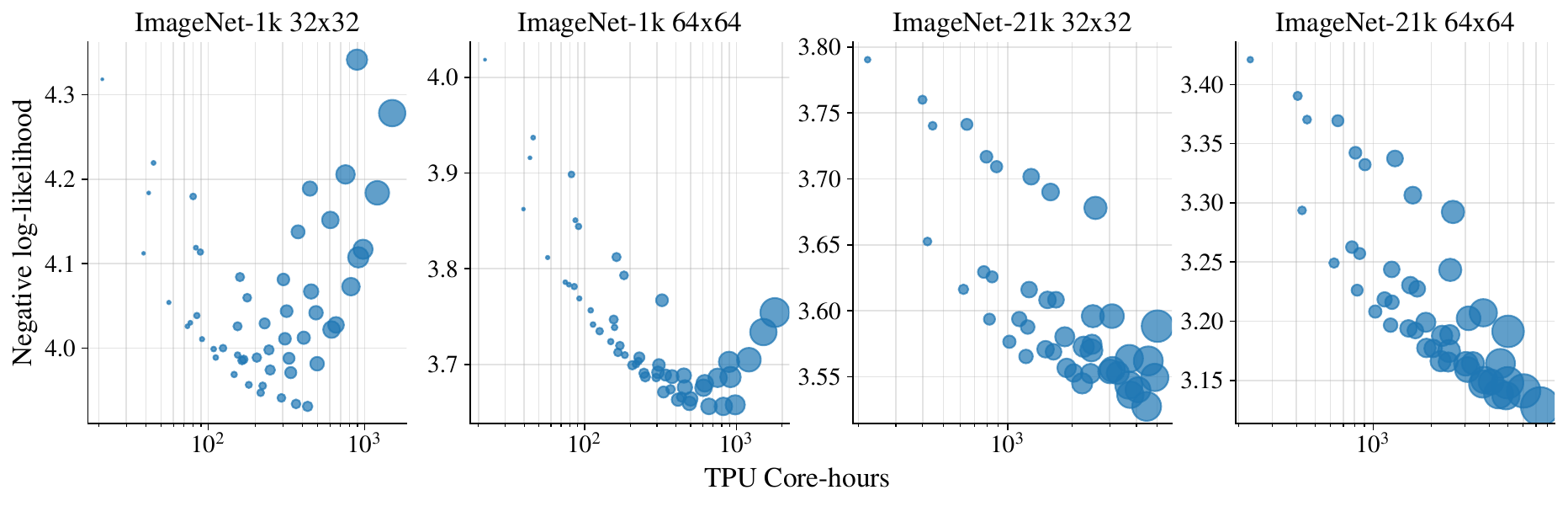}
    \caption{NLL as a function of training compute obtained when training Jet architectures with a range of architecture hyper-paramaters, for 4 different data sets. The size of each marker is proportional to the number of parameters in the model configuration. Overall we observe normalizing flow models benefit from scale, yet ImageNet-1k models start to overfit. When increasing the amount of data to ImageNet-21k size, we observe little overfitting and strong scaling trends.}
    \label{fig:core_sweeps}
\end{figure}

\begin{table}[t]
    \caption{Negative log-likelihood of Jet models and the state-of-the-art coupling-based normalizing flow model Flow++~\citep{ho2019flow++} from the literature.
    Jet (21k) indicates the result obtained when using a Jet model pretrained on ImageNet21k.
    Overall, we observe that all tasks benefit from this pretraining and that Jet matches or outperforms Flow++.}
    \label{tab:results}
    \centering

    \begin{subtable}[t]{0.32\textwidth}
        \caption{ImageNet-1k $64\times64$}
        \centering
        \begin{tabular}{lc}
            \toprule
            Model & Result (NLL$\downarrow$) \\
            \midrule
            Flow++ & 3.69\phantom{0} \\
            Jet & 3.656 \\
            Jet (I21k) & 3.580 \\
            \bottomrule
        \end{tabular}
        \label{tab:results-task1}
    \end{subtable}
    \hfill
    \begin{subtable}[t]{0.32\textwidth}
        \caption{ImageNet-1k $32\times32$}
        \centering
        \begin{tabular}{lc}
            \toprule
            Model & Result (NLL$\downarrow$) \\
            \midrule
            Flow++ & 3.86\phantom{0} \\
            Jet & 3.931 \\
            Jet (I21k) & 3.857 \\
            \bottomrule
        \end{tabular}
        \label{tab:results-task2}
    \end{subtable}
    \hfill
    \begin{subtable}[t]{0.32\textwidth}
        \caption{CIFAR-10 $32\times32$}
        \centering
        \begin{tabular}{lc}
            \toprule
            Model & Result (NLL$\downarrow$) \\
            \midrule
            Flow++ & 3.08\phantom{0} \\
            Jet (I21k) & 3.018 \\
            & \\
            \bottomrule
        \end{tabular}
        \label{tab:results-task3}
    \end{subtable}
\end{table}

We conduct extensive sweep that includes Jet models trained across varying computational capacity, data size (ImageNet-1k and ImageNet-21k) and resolutions ($32\times32$ and $64\times64$). For the model capacity sweep we explore the following configurations (approximately spanning 2 orders of magnitude in compute intensity):
\begin{itemize}
  \item the number of coupling layers in $\{16, 32, 64\}$ for ImageNet-1k and in $\{32, 64, 128\}$ for ImageNet-21k,
  \item the depth of ViT used inside the coupling layer in $\{1, 2, 3, 4, 5, 6\}$,
  \item the ViT embedding dimension in $\{256, 512, 768\}$ for ImageNet-1k and in $\{512, 768, 1024\}$ for ImageNet-21k. The number of heads is tied to the embedding dimension and equal to $\{4, 8, 12\}$ and $\{8, 12, 16\}$  respectively.
\end{itemize}
We use a fixed standard learning rate of $3$e$-4$, weight decay of $1$e$-5$ and train for 200 epochs for ImageNet-1k and for 50 epochs for ImageNet-21k. We additionally investigate transfer learning setup and finetune our best ImageNet-21k models on ImageNet-1k and CIFAR-10. Full sweep results are shown in Figure~\ref{fig:core_sweeps}. Additionally, we present key numerical results in Table~\ref{tab:results}.

Our first observation is that due to high expressive power of the Jet model parameterized by a ViT model, it tends to quickly overfit on ImageNet-1k, which has only 1.2M examples in total. However, despite this, for $64 \times 64$ Jet attains a state-of-the-art NLL of $3.66$ bpd. For $32\times32$ input resolution overfitting for large models is more severe, however reasonably small models still achieve a competitive result of $3.93$ bpd.

It is important to note that unlabeled natural images are abundant. Thus, the best way to tackle overfitting is to increase the amount of training data, as opposed to constraining or regularizing the model. We, therefore, use the Imagenet-21k dataset with more than $10\times$ more images than in ImageNet-1k. As expected, overfitting is tamed, with larger models leading to increasingly better results.    

The next important question is whether a model trained on a larger and more class-diverse ImageNet-21k dataset transfers to ImageNet-1k. We observe that with a very light finetuning (30 epochs, learning rate of $1$e$-5$ and $3$e$-6$ for higher resolution) we obtain state-of-the-art results on ImageNet-1k, attaining $3.58$ and $3.86$ bpd on $32\times32$ and $64\times64$ input resolution, respectively. However, one can argue that ImageNet-1k is very similar to ImageNet-21k and results are overly optimistic. To address this concern, we also transfer the ImageNet-21k model to the CIFAR-10 dataset, which has very different type of images and has a different downsampling procedure. Again, after finetuning (100 epochs, learning rate of $3$e$-6$) we obtain state-of-the-art performance of $3.02$ bpd.

Overall, our results provide a strong evidence that the Jet architecture achieves state-of-the art performance in the class of coupling-based normalizing flow models, and, at the same time, it is very stable and easy to train across wide range of scenarios.

\subsection{Ablations}

In this section we ablate key design choices for the Jet model.
Our default ablation setting is moderately-sized Jet model, trained on ImageNet-1k $64\times64$ for 200 epochs.
We set the total depth to 32 coupling layers, the ViT depth to 2 blocks and width to 512 dimensions.
In our scaling study this configuration was reasonably close to the optimal setup, while being sufficiently fast for the extensive ablations sweeps. 
We use negative log-likelihood reported in bpd as the main ablation metric.

\subsubsection{Coupling types}
\label{sec:coupling-types}

\newcommand{\gc}[1]{\textcolor{darkgray}{#1}}  

\begin{figure}[t]
\centering
\includegraphics[width=1\textwidth]{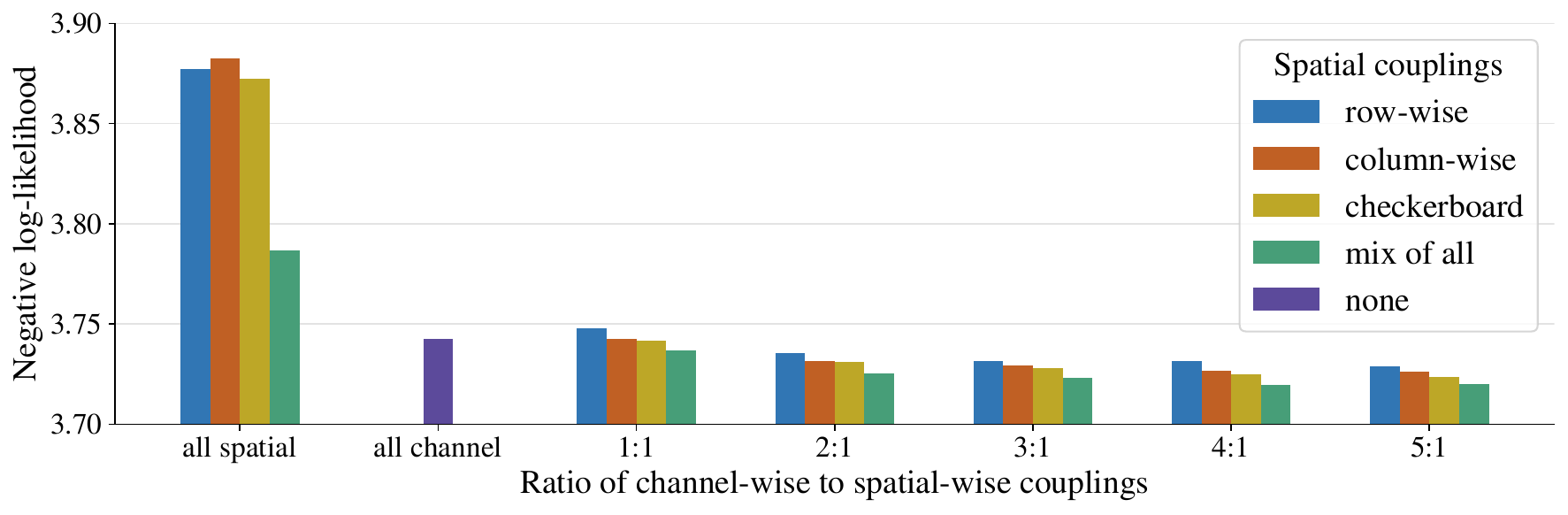}
\caption{Ablation of coupling types. Negative log-likelihood on ImageNet-1k $64$x$64$ when varying the ratio of channel-wise to spatial-wise couplings and when using different types of spatial-wise couplings. Results in table format in Appendix Table~\ref{tab:coupling-kinds}.}
\label{fig:coupling-kinds}
\end{figure}

As presented in Section~\ref{sec:dimension-split}, we consider 4 different ways to split the channels in a coupling block: one is a channel-wise method (splits the channels into two parts) and three patch-wise methods (alternating rows, alternating columns or ``checkerboard'' splitting).
Inspired by the prior literature, where channel-wise couplings were more prominent, we explore the following design space: $M$ repeated channel-wise couplings followed by $1$ spatial coupling. The spatial coupling is either fixed to one of the methods, or alternates between 3 types. We vary $M$ from 0 to 5. Overall, the above choice space gives rise to 25 configurations.

Figure~\ref{fig:coupling-kinds} presents the quantitative results, suggesting the following insights:
\begin{itemize}

\item Using interleaved spatial and channel coupling is optimal, where $M=4$ is optimal, though $M \in {2..5}$ achieves very similar performance. 
\item Alternating spatial coupling methods are superior over using any single coupling types.
\item Spatial coupling alone is the worst configuration, while exclusive channel coupling is competitive.  
\end{itemize}

\subsubsection{Coupling layers vs ViT depth}

Jet has two depth parameters: number of coupling layers and ViT depth within each coupling layer. The total compute is roughly proportional to the product of these two depths. We observe a very interesting interplay between these parameters, see Figure~\ref{fig:block-depth} representing ImageNet-21k $32\times32$, which is a detailed view of Figure~\ref{fig:core_sweeps}.

Specifically, we observe that scaling the number of coupling layers, while keeping shallow ViT models (e.g. depth 1) results in an unfavorable compute-performance trade-off. It appears that ViT depth of at least $4$--$6$ is the necessary condition for the Jet model to stay close to the frontier.  For example, a model with 32 coupling layers and ViT depth 4 has roughly the same compute requirements as a model with 128 coupling layers and ViT depth 1. However, the former performs much better than the latter for fixed compute. 

\subsubsection{ViT vs CNNs}

\begin{table}[t]
\caption{Ablation of design choices reporting negative log-likelihood (bpd) on ImageNet-1k $64\times64$.
Table~\ref{tab:mask-vs-pair}: shows the effect of using masking-mode or pairing-mode (best)when splitting the spatial tokens.
Table~\ref{tab:act-norm-invertible-dense}: shows the effect of using invertible dense layers and/or activation norm. Best results obtained when not using either.
}
\centering
\subfloat[][]{
\begin{tabular}{crr}
\toprule
Couplings ratio & Masking & Pairing \\
\midrule
all spatial &  \gc{3.844} & 3.787 \\
1:1 & \gc{3.741} & 3.737 \\
2:1 & \gc{3.727} & 3.725 \\
3:1 & 3.723 & 3.723 \\
4:1 & \gc{3.722} & 3.719 \\
5:1 & \gc{3.722} & 3.720 \\
\bottomrule
\end{tabular}
\label{tab:mask-vs-pair}
}
\qquad
\subfloat[][]{
\begin{tabular}{ccl}
\toprule
Activation & Invertible & NLL$\downarrow$ \\
Norm & Dense & \\
\midrule
$\times$ & $\times$ & \underline{3.720} \\
\checkmark & $\times$ & 3.727 \\
$\times$& \checkmark & 3.741 \\
\checkmark & \checkmark & 3.733 \\
\bottomrule
\end{tabular}
\label{tab:act-norm-invertible-dense}
}
\end{table}

To ablate the use a ViT instead of a CNN block, we conduct a similar sweep to our main sweep on ImageNet-1k $64\times64$ but using a CNN architecture (specifically we use the CNN architrecture from~\citep{kolesnikov2020big}). This time sweeping the following settings for the CNN setup: model depth in $\{16, 32, 64\}$, CNN block depth in $\{1, \ldots, 8\}$, block embedding dimension in $\{256, 512, 768, 1024, 1536\}$. Block dimension 1536 was not used for model depth 64 due to significant memory costs. The results in Figure~\ref{fig:ablate-vit-cnn} show that the CNN-based variant lags significantly behind the ViT-based one.

We anticipate that gap can be reduced by using multiscale architectures, as commonly done in the literature~\citep{dinh2017realnvp,kingma2018glow}. However, in this paper we strive to simplify design and exclude multiscale architectures.

\subsubsection{Coupling implementation}

We investigate two common approaches for implementing coupling layers: masking or pairing.
To be concrete, let's assume that we implement a coupling layer that splits the input spatially into two gropus of patches.
In one approach, which we name ``masking'' mode, we feed the $K$ patches to the ViT block but mask with zeros the ones corresponding to the $x_2$ group. At the output we use only the ones corresponding to the $x_2$ group and ignore the output of the patches corresponding to the $x_1$ group.
One potential issue with this method is that it weakens the residual connections as the tokens from which we predict the output are tokens which observe zero as input.

Another approach we consider is a ``pairing'' mode in which we establish a pairing between input and output patches (or embeddings).
For example when using a vertical-stripes pattern, the outputs of the ViT block for a patch in the $x_1$ group will predict the scale and bias for a patch in the $x_2$ group (e.g. to the patch below). This would make the ViT block processes only $N/2$ patches.

We experiment with these two implementation types while sweeping the $M$:1 channel:spatial coupling ratios as in Section~\ref{sec:coupling-types}. The results presented in Table~\ref{tab:mask-vs-pair} indicate pairing to be superior, though the impact becomes smaller as one increases the number of channel couplings which do not depend on this design decision. We observe that both methods perform very similarly, with the pairing being slightly ahead of masking, especially for the scenario when only spatial couplings are used. Thus, for the Jet model, we default to using pairing mode.

\subsubsection{Invertible dense layers and activation normalization}
Glow~\citep{kingma2018glow} introduces two components to improve the performance of normalizing flows:
(1) a learnable, invertible dense layer which replaces the fixed permutation used to split the channels for each coupling;
(2) an activation normalization layer with a scalar and bias parameters per channel similar to batch normalization.

We use neither of them in Jet, but ablate whether performance could be improved by introducing those in Table~\ref{tab:act-norm-invertible-dense}.
As a note, we observed that the use of activation normalization alone to be highly unstable, we sweep additional learning rate and seeds and report the best result found.
Overall we found that not using any of those components leads to the best results.

\subsubsection{Uniform dequantization vs dequantization flow}
Flow++~\citep{ho2019flow++} introduces a variational dequantization scheme to normalizing flows. Concretely, it proposes to replace the uniform dequantization noise added to the input with an image-conditional, learned noise distribution modeled by another normalizing flow. We ablate this component by training a 64 layer Jet model with 16-layer dequantization flow and compare it with an 80-layer base Jet model. The image conditioning of the dequantization flow was implemented by adding cross-attention layers to the ViT-blocks to the input. 
We observe no significant improvements when using the dequantization flow component.

\section{Related work}

NICE~\citep{dinh2014nice} popularized coupling-based normalizing flows with the introduction of the additive coupling layer. RealNVP~\citep{dinh2017realnvp} then increased the flow's expressivity by using affine coupling layers in combination with a multiscale architecture, and \citep{kingma2018glow, hoogeboom2019emerging, sukthanker2022generative} proposed additional specialized invertible layers for image modeling. Flow++
~\citep{ho2019flow++} demonstrated improvements from learning the dequantization noise distribution along with the flow model.

Another class of likelihood-based generative models are autoregressive models which flatten the (sub)pixels of an image into a sequence. Autoregressive modeling is enabled by using CNNs~\citep{van2016pixel, van2016conditional, salimans2016pixelcnn} or transformers~\citep{parmar2018image,chen2020generative}. \cite{kolesnikov2017pixelcnn, menick2018generating} improved performance of autoregressive models with hierarchical modeling (e.g. over color depth or resolution). While obtaining better results than normalizing flows, autoregressive models are also much slower and do not scale to large resolutions as they require a forward-pass per (sub)pixel.

In the context of normalizing flows, autoregressive dependency patterns between latent variables are a popular approach to improve modeling capabilities of normalizing flows~\citep{kingma2016improved, papamakarios2017masked, huang2018neural}. \cite{bhattacharyya2020normalizing} combined autoregressive modeling with a multiscale architecture. Concurrently to this work, \cite{zhai2024normalizing} proposed a combination of the transformer-based autoregressive flow.

\section{Conclusion}

The Jet model revisits normalizing flows with a focus on simplicity and performance. While eliminating complex components such as multiscale architectures and invertible layers, Jet achieves state-of-the-art results across benchmarks while maintaining a straightforward design.

We see normalizing flows, and Jet in particular, as a useful tool for advancing generative modeling. Due to its simple structure and lossless guarantees, it can serve as a building block for powerful generative systems. One recent example is \citep{tschannen2024jetformer}, which leverages a normalizing flow to enable end-to-end autoregressive modeling of raw high-resolution images. We anticipate more progress in this area and believe that the Jet model will prove itself a powerful normalizing flow component that can be used out-of-the-box for a variety of applications.

\paragraph{Acknowledgments.} We thank Fabian Mentzer for discussions and feedback on this paper.

\clearpage

\bibliography{main}

\begin{thebibliography}{26}
\providecommand{\natexlab}[1]{#1}
\providecommand{\url}[1]{\texttt{#1}}
\expandafter\ifx\csname urlstyle\endcsname\relax
  \providecommand{\doi}[1]{doi: #1}\else
  \providecommand{\doi}{doi: \begingroup \urlstyle{rm}\Url}\fi

\bibitem[Bhattacharyya et~al.(2020)Bhattacharyya, Mahajan, Fritz, Schiele, and
  Roth]{bhattacharyya2020normalizing}
Apratim Bhattacharyya, Shweta Mahajan, Mario Fritz, Bernt Schiele, and Stefan
  Roth.
\newblock Normalizing flows with multi-scale autoregressive priors.
\newblock In \emph{CVPR}, 2020.

\bibitem[Chen et~al.(2020)Chen, Radford, Child, Wu, Jun, Luan, and
  Sutskever]{chen2020generative}
Mark Chen, Alec Radford, Rewon Child, Jeffrey Wu, Heewoo Jun, David Luan, and
  Ilya Sutskever.
\newblock Generative pretraining from pixels.
\newblock In \emph{ICML}, 2020.

\bibitem[Chen et~al.(2016)Chen, Kingma, Salimans, Duan, Dhariwal, Schulman,
  Sutskever, and Abbeel]{chen2016variational}
Xi~Chen, Diederik~P Kingma, Tim Salimans, Yan Duan, Prafulla Dhariwal, John
  Schulman, Ilya Sutskever, and Pieter Abbeel.
\newblock Variational lossy autoencoder.
\newblock \emph{arXiv:1611.02731}, 2016.

\bibitem[Chrabaszcz et~al.(2017)Chrabaszcz, Loshchilov, and
  Hutter]{chrabaszcz2017downsampled}
Patryk Chrabaszcz, Ilya Loshchilov, and Frank Hutter.
\newblock A downsampled variant of imagenet as an alternative to the cifar
  datasets.
\newblock \emph{arXiv:1707.08819}, 2017.

\bibitem[Dinh et~al.(2014)Dinh, Krueger, and Bengio]{dinh2014nice}
Laurent Dinh, David Krueger, and Yoshua Bengio.
\newblock Nice: Non-linear independent components estimation.
\newblock \emph{arXiv:1410.8516}, 2014.

\bibitem[Dinh et~al.(2017)Dinh, Sohl-Dickstein, and Bengio]{dinh2017realnvp}
Laurent Dinh, Jascha Sohl-Dickstein, and Samy Bengio.
\newblock Density estimation using real {NVP}.
\newblock In \emph{ICLR}, 2017.

\bibitem[Dosovitskiy et~al.(2021)Dosovitskiy, Beyer, Kolesnikov, Weissenborn,
  Zhai, Unterthiner, Dehghani, Minderer, Heigold, Gelly, Uszkoreit, and
  Houlsby]{dosovitskiy2020image}
Alexey Dosovitskiy, Lucas Beyer, Alexander Kolesnikov, Dirk Weissenborn,
  Xiaohua Zhai, Thomas Unterthiner, Mostafa Dehghani, Matthias Minderer, Georg
  Heigold, Sylvain Gelly, Jakob Uszkoreit, and Neil Houlsby.
\newblock An image is worth 16x16 words: Transformers for image recognition at
  scale.
\newblock \emph{ICLR}, 2021.

\bibitem[Ho et~al.(2019)Ho, Chen, Srinivas, Duan, and Abbeel]{ho2019flow++}
Jonathan Ho, Xi~Chen, Aravind Srinivas, Yan Duan, and Pieter Abbeel.
\newblock Flow++: Improving flow-based generative models with variational
  dequantization and architecture design.
\newblock In \emph{ICML}, 2019.

\bibitem[Hoogeboom et~al.(2019)Hoogeboom, Van Den~Berg, and
  Welling]{hoogeboom2019emerging}
Emiel Hoogeboom, Rianne Van Den~Berg, and Max Welling.
\newblock Emerging convolutions for generative normalizing flows.
\newblock In \emph{ICML}, 2019.

\bibitem[Huang et~al.(2018)Huang, Krueger, Lacoste, and
  Courville]{huang2018neural}
Chin-Wei Huang, David Krueger, Alexandre Lacoste, and Aaron Courville.
\newblock Neural autoregressive flows.
\newblock In \emph{ICML}, 2018.

\bibitem[Kingma \& Dhariwal(2018)Kingma and Dhariwal]{kingma2018glow}
Durk~P Kingma and Prafulla Dhariwal.
\newblock Glow: Generative flow with invertible 1x1 convolutions.
\newblock \emph{NeurIPS}, 2018.

\bibitem[Kingma et~al.(2016)Kingma, Salimans, Jozefowicz, Chen, Sutskever, and
  Welling]{kingma2016improved}
Durk~P Kingma, Tim Salimans, Rafal Jozefowicz, Xi~Chen, Ilya Sutskever, and Max
  Welling.
\newblock Improved variational inference with inverse autoregressive flow.
\newblock \emph{NeurIPS}, 2016.

\bibitem[Kolesnikov \& Lampert(2017)Kolesnikov and
  Lampert]{kolesnikov2017pixelcnn}
Alexander Kolesnikov and Christoph~H Lampert.
\newblock Pixel{CNN} models with auxiliary variables for natural image
  modeling.
\newblock In \emph{ICML}, 2017.

\bibitem[Kolesnikov et~al.(2020)Kolesnikov, Beyer, Zhai, Puigcerver, Yung,
  Gelly, and Houlsby]{kolesnikov2020big}
Alexander Kolesnikov, Lucas Beyer, Xiaohua Zhai, Joan Puigcerver, Jessica Yung,
  Sylvain Gelly, and Neil Houlsby.
\newblock Big transfer {(BiT)}: General visual representation learning.
\newblock In \emph{ECCV}, 2020.

\bibitem[Loshchilov et~al.(2017)Loshchilov, Hutter,
  et~al.]{loshchilov2017fixing}
Ilya Loshchilov, Frank Hutter, et~al.
\newblock Fixing weight decay regularization in adam.
\newblock \emph{arXiv:1711.05101}, 2017.

\bibitem[Menick \& Kalchbrenner(2019)Menick and
  Kalchbrenner]{menick2018generating}
Jacob Menick and Nal Kalchbrenner.
\newblock Generating high fidelity images with subscale pixel networks and
  multidimensional upscaling.
\newblock \emph{ICLR}, 2019.

\bibitem[Papamakarios et~al.(2017)Papamakarios, Pavlakou, and
  Murray]{papamakarios2017masked}
George Papamakarios, Theo Pavlakou, and Iain Murray.
\newblock Masked autoregressive flow for density estimation.
\newblock \emph{NeurIPS}, 2017.

\bibitem[Parmar et~al.(2018)Parmar, Vaswani, Uszkoreit, Kaiser, Shazeer, Ku,
  and Tran]{parmar2018image}
Niki Parmar, Ashish Vaswani, Jakob Uszkoreit, Lukasz Kaiser, Noam Shazeer,
  Alexander Ku, and Dustin Tran.
\newblock Image transformer.
\newblock In \emph{ICML}, 2018.

\bibitem[Salimans et~al.(2016)Salimans, Karpathy, Chen, and
  Kingma]{salimans2016pixelcnn}
Tim Salimans, Andrej Karpathy, Xi~Chen, and Diederik~P Kingma.
\newblock {PixelCNN++}: Improving the {PixelCNN} with discretized logistic
  mixture likelihood and other modifications.
\newblock In \emph{ICLR}, 2016.

\bibitem[Sukthanker et~al.(2022)Sukthanker, Huang, Kumar, Timofte, and
  Van~Gool]{sukthanker2022generative}
Rhea~Sanjay Sukthanker, Zhiwu Huang, Suryansh Kumar, Radu Timofte, and Luc
  Van~Gool.
\newblock Generative flows with invertible attentions.
\newblock In \emph{CVPR}, 2022.

\bibitem[Theis et~al.(2015)Theis, Oord, and Bethge]{theis2015note}
Lucas Theis, A{\"a}ron van~den Oord, and Matthias Bethge.
\newblock A note on the evaluation of generative models.
\newblock \emph{arXiv:1511.01844}, 2015.

\bibitem[Tschannen et~al.(2024{\natexlab{a}})Tschannen, Eastwood, and
  Mentzer]{tschannen2023givt}
Michael Tschannen, Cian Eastwood, and Fabian Mentzer.
\newblock {GIVT}: Generative infinite-vocabulary transformers.
\newblock In \emph{ECCV}, 2024{\natexlab{a}}.

\bibitem[Tschannen et~al.(2024{\natexlab{b}})Tschannen, Pinto, and
  Kolesnikov]{tschannen2024jetformer}
Michael Tschannen, Andr{\'e}~Susano Pinto, and Alexander Kolesnikov.
\newblock {JetFormer}: An autoregressive generative model of raw images and
  text.
\newblock \emph{arXiv:2411.19722}, 2024{\natexlab{b}}.

\bibitem[Van~den Oord et~al.(2016{\natexlab{a}})Van~den Oord, Kalchbrenner,
  Espeholt, Vinyals, and Graves]{van2016conditional}
Aaron Van~den Oord, Nal Kalchbrenner, Lasse Espeholt, Oriol Vinyals, and Alex
  Graves.
\newblock Conditional image generation with {PixelCNN} decoders.
\newblock \emph{NeurIPS}, 2016{\natexlab{a}}.

\bibitem[Van~den Oord et~al.(2016{\natexlab{b}})Van~den Oord, Kalchbrenner, and
  Kavukcuoglu]{van2016pixel}
A{\"a}ron Van~den Oord, Nal Kalchbrenner, and Koray Kavukcuoglu.
\newblock Pixel recurrent neural networks.
\newblock In \emph{ICML}, 2016{\natexlab{b}}.

\bibitem[Zhai et~al.(2024)Zhai, Zhang, Nakkiran, Berthelot, Gu, Zheng, Chen,
  Bautista, Jaitly, and Susskind]{zhai2024normalizing}
Shuangfei Zhai, Ruixiang Zhang, Preetum Nakkiran, David Berthelot, Jiatao Gu,
  Huangjie Zheng, Tianrong Chen, Miguel~Angel Bautista, Navdeep Jaitly, and
  Josh Susskind.
\newblock Normalizing flows are capable generative models.
\newblock \emph{arXiv:2412.06329}, 2024.

\end{thebibliography}
\bibliographystyle{tmlr}

\clearpage
\appendix
\section{Appendix}

\subsection{Architecture details}

\begin{table}[h]
\caption{Architecture details for models used to obtain the main results in Table~\ref{tab:results}.}
\centering
\begin{tabular}{lrrrr}
\toprule
  & \multicolumn{2}{c}{ImageNet-1k} & \multicolumn{2}{c}{ImageNet-21k} \\
  & $32\times32$ & $64\times64$ & $32\times32$ & $64\times64$ \\
\midrule
Coupling layers          &    64  &   64 &   64 &    64 \\
ViT depth                &     6  &    5 &    6 &     6 \\
ViT width                &    256 &  512 &  768 &  1024 \\
ViT attention heads      &      4 &    8 &   12 &    16 \\
\bottomrule
\end{tabular}
\end{table}

\subsection{Detailed results}

\begin{table}[h]
\caption{NLL on ImageNet-1k $32\times32$ when sweeping architecture hyper-parameters.}
\centering
\begin{tabular}{lrrr|rrr|rrr}
\toprule
depth & \multicolumn{3}{c|}{16} & \multicolumn{3}{c|}{32} & \multicolumn{3}{c}{64} \\
ViT dim & 256 & 512 & 768 & 256 & 512 & 768 & 256 & 512 & 768 \\
\midrule
ViT depth &  &  &  &  &  &  &  &  &  \\
1 & 4.32 & 4.22 & 4.18 & 4.18 & 4.11 & 4.08 & 4.12 & 4.06 & 4.04 \\
2 & 4.11 & 4.04 & 4.03 & 4.03 & 3.99 & 4.01 & 3.99 & 3.97 & 4.02 \\
3 & 4.05 & 4.00 & 4.03 & 3.99 & 3.97 & 4.07 & 3.96 & 3.98 & 4.11 \\
4 & 4.03 & 3.99 & 4.08 & 3.97 & 3.99 & 4.15 & 3.94 & 4.03 & 4.18 \\
5 & 4.01 & 3.99 & 4.14 & 3.96 & 4.01 & 4.21 & 3.93 & 4.07 & 4.28 \\
6 & 4.00 & 4.00 & 4.19 & 3.95 & 4.04 & 4.34 & 3.93 & 4.12 & N/A \\
\bottomrule
\end{tabular}
\end{table}

\begin{table}[h]
\caption{NLL on ImageNet-1k $64\times64$ when sweeping architecture hyper-parameters.}
\centering
\begin{tabular}{lrrr|rrr|rrr}
\toprule
Couplings & \multicolumn{3}{c|}{16} & \multicolumn{3}{c|}{32} & \multicolumn{3}{c}{64} \\
ViT dim & 256 & 512 & 768 & 256 & 512 & 768 & 256 & 512 & 768 \\
\midrule
ViT depth &  &  &  &  &  &  &  &  &  \\
1 & 4.02 & 3.94 & 3.90 & 3.92 & 3.84 & 3.81 & 3.85 & 3.79 & 3.77 \\
2 & 3.86 & 3.78 & 3.75 & 3.78 & 3.72 & 3.70 & 3.74 & 3.69 & 3.68 \\
3 & 3.81 & 3.73 & 3.71 & 3.74 & 3.69 & 3.68 & 3.70 & 3.66 & 3.69 \\
4 & 3.79 & 3.71 & 3.69 & 3.72 & 3.67 & 3.68 & 3.69 & 3.66 & 3.70 \\
5 & 3.77 & 3.70 & 3.69 & 3.71 & 3.66 & 3.69 & 3.67 & 3.66 & 3.73 \\
6 & 3.76 & 3.69 & 3.69 & 3.70 & 3.66 & 3.70 & 3.67 & 3.66 & 3.75 \\
\bottomrule
\end{tabular}
\end{table}

\begin{table}[h]
\caption{NLL on ImageNet-1k $64\times64$ when sweeping CNN architecture hyper-parameters.}
\centering

\begin{tabular}{lrrrrr|rrrrr|rrrr}
\toprule
Couplings & \multicolumn{5}{c|}{16} & \multicolumn{5}{c|}{32} & \multicolumn{4}{c}{64} \\
Block dim & 256 & 512 & 768 & 1024 & 1536 & 256 & 512 & 768 & 1024 & 1536 & 256 & 512 & 768 & 1024 \\
\midrule
\multicolumn{2}{l}{CNN depth} &  &  &  &  &  &  &  &  &  &  &  &  &   \\
1 & 4.17 & 4.05 & 4.00 & 3.96 & 3.92 & 4.05 & 3.95 & 3.91 & 3.87 & 3.83 & 3.97 & 3.88 & 3.84 & N/A \\
2 & 4.09 & 3.97 & N/A & 3.87 & 3.83 & 3.98 & 3.88 & 3.83 & 3.79 & 3.76 & 3.90 & 3.81 & 3.77 & 3.75 \\
3 & 4.06 & 3.94 & 3.88 & 3.84 & 3.80 & 3.95 & 3.84 & 3.79 & 3.77 & 3.74 & 3.87 & 3.78 & 3.75 & 3.73 \\
4 & 4.04 & 3.91 & 3.85 & 3.82 & 3.78 & 3.93 & 3.83 & 3.78 & 3.75 & 3.74 & 3.85 & 3.76 & 5.97 & N/A \\
5 & 4.02 & 3.90 & 3.84 & 3.81 & 3.77 & 3.91 & 3.81 & 3.77 & 3.75 & 3.74 & 3.84 & 3.76 & 3.98 & N/A \\
6 & 4.00 & 3.88 & 3.83 & 3.80 & 3.77 & 3.90 & 3.80 & 3.76 & 3.74 & 3.74 & 3.83 & 3.89 & 6.41 & 3.77 \\
8 & 3.98 & 3.87 & 3.81 & 3.78 & 3.77 & 3.88 & 3.79 & 3.75 & 3.74 & 3.75 & 3.81 & 3.74 & 3.87 & 5.89 \\
\bottomrule
\end{tabular}
\end{table}

\begin{table}[h]
\caption{NLL on ImageNet-21k $32\times32$ when sweeping architecture hyper-parameters.}
\centering
\begin{tabular}{lrrr|rrr|rrr}
\toprule
Couplings & \multicolumn{3}{c|}{32} & \multicolumn{3}{c|}{64} & \multicolumn{3}{c}{128} \\
ViT dim & 512 & 768 & 1024 & 512 & 768 & 1024 & 512 & 768 & 1024 \\
\midrule
ViT depth &  &  &  &  &  &  &  &  &  \\
1 & 3.79 & 3.76 & 3.74 & 3.74 & 3.72 & 3.70 & 3.71 & 3.69 & 3.68 \\
2 & 3.65 & 3.63 & 3.62 & 3.63 & 3.61 & 3.60 & 3.61 & 3.60 & 3.59 \\
3 & 3.62 & 3.59 & 3.58 & 3.59 & 3.57 & 3.56 & 3.57 & 3.56 & N/A \\
4 & 3.59 & 3.57 & 3.57 & 3.57 & 3.55 & N/A & 3.55 & N/A & N/A \\
5 & 3.58 & 3.56 & 3.56 & 3.55 & 3.54 & N/A & 3.54 & N/A & N/A \\
6 & 3.57 & 3.55 & 3.54 & 3.55 & 3.53 & N/A & 3.55 & N/A & N/A \\
\bottomrule
\end{tabular}
\end{table}

\begin{table}[h]
\caption{NLL on ImageNet-21k $64\times64$ when sweeping architecture hyper-parameters.}
\centering
\begin{tabular}{lrrr|rrr|rrr}
\toprule
Couplings & \multicolumn{3}{c|}{32} & \multicolumn{3}{c|}{64} & \multicolumn{3}{c}{128} \\
ViT dim & 512 & 768 & 1024 & 512 & 768 & 1024 & 512 & 768 & 1024 \\
\midrule
ViT depth &  &  &  &  &  &  &  &  &  \\
1 & 3.42 & 3.39 & 3.37 & 3.37 & 3.34 & 3.34 & 3.33 & 3.31 & 3.29 \\
2 & 3.29 & 3.26 & 3.24 & 3.26 & 3.23 & 3.24 & 3.23 & 3.20 & 3.19 \\
3 & 3.25 & 3.22 & 3.20 & 3.22 & 3.19 & 3.21 & 3.19 & 3.16 & N/A \\
4 & 3.23 & 3.19 & 3.17 & 3.19 & 3.16 & 3.15 & 3.16 & 3.14 & N/A \\
5 & 3.21 & 3.18 & 3.16 & 3.18 & 3.15 & N/A & 3.15 & N/A & N/A \\
6 & 3.20 & 3.17 & 3.15 & 3.17 & 3.14 & 3.13 & 3.14 & N/A & N/A \\
\bottomrule
\end{tabular}
\end{table}

\begin{table}[h]
\caption{NLL on ImageNet-1k $64\times64$ when varying the ratio between channel and spatial couplings.}
\centering
\begin{tabular}{crrrr}
\toprule
Couplings ratio & \multicolumn{4}{c}{Type of spatial couplings} \\
(channel : spatial)  & \multicolumn{1}{c}{row-wise} & \multicolumn{1}{c}{column-wise} & \multicolumn{1}{c}{checkerboard} & \multicolumn{1}{c}{mix of all} \\
\midrule
all spatial           & \gc{3.877} & \gc{3.882} & \gc{3.872} & {3.787} \\
1:1 & \gc{3.747} & \gc{3.742} & \gc{3.741} & {3.737} \\
2:1 & \gc{3.735} & \gc{3.731} & \gc{3.731} & {3.725} \\
3:1 & \gc{3.732} & \gc{3.729} & \gc{3.728} &  {3.723} \\
4:1 & \gc{3.731} & \gc{3.726} & \gc{3.724} & \underline{3.719} \\
5:1 & \gc{3.728} & \gc{3.726} & \gc{3.723} & {3.720} \\
\midrule
all channel & \multicolumn{4}{c}{3.742} \\
\bottomrule
\end{tabular}
\label{tab:coupling-kinds}
\end{table}

\begin{figure}[ht]
\centering

    \begin{subfigure}[c]{0.45\columnwidth}
        \includegraphics[width=1\textwidth]{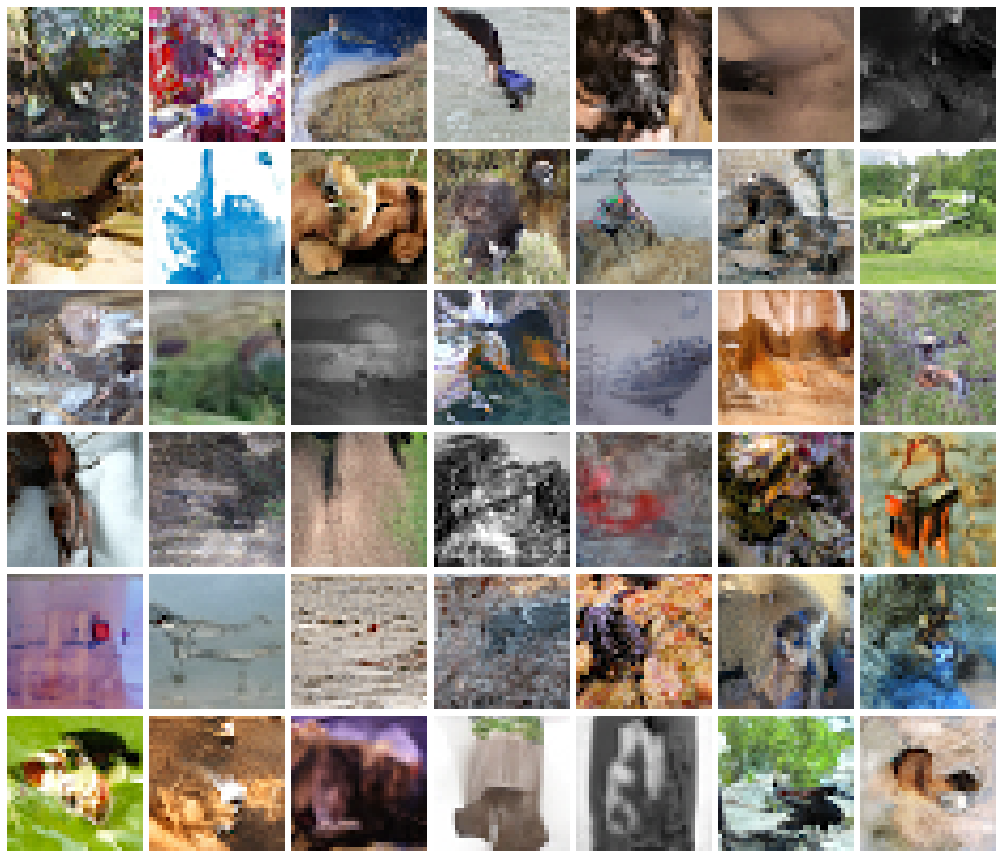}\caption{Jet -- $32\times32$}
    \end{subfigure}
    \hspace{0.01\columnwidth}
    \begin{subfigure}[c]{0.45\columnwidth}
        \includegraphics[width=1\textwidth]{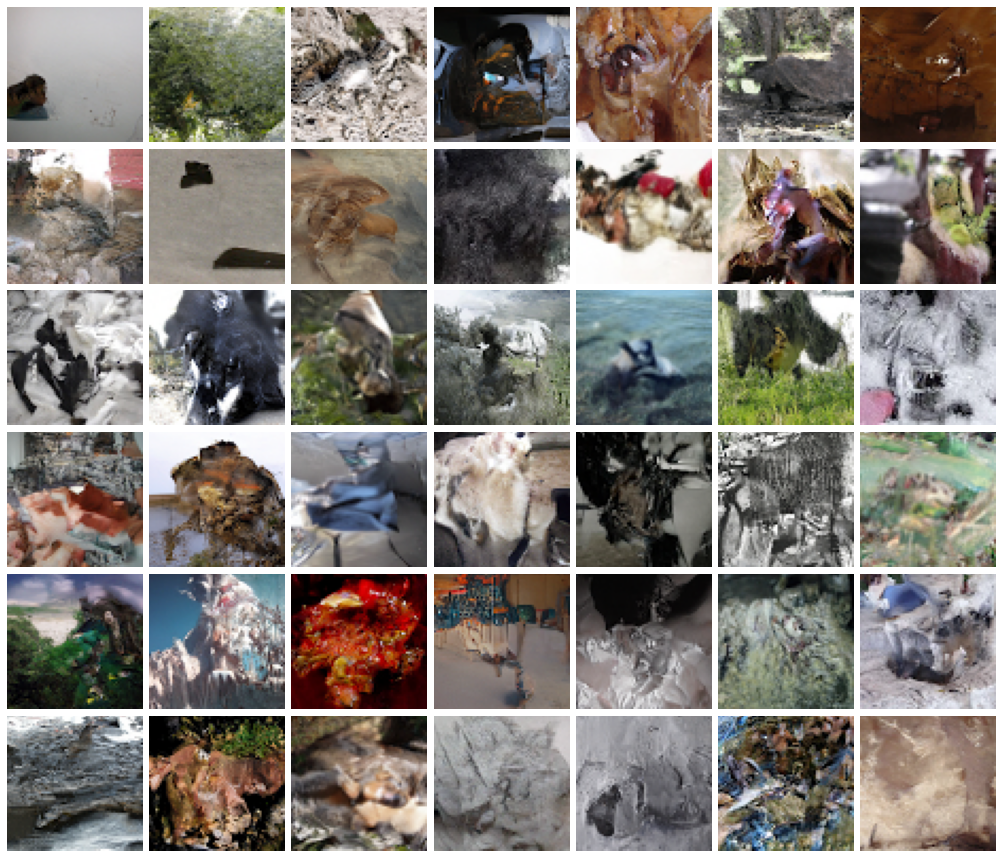}\caption{Jet  -- $64\times64$}
    \end{subfigure}
    \\
    \vspace{0.01\columnwidth}

    \begin{subfigure}[c]{0.45\columnwidth}
        \includegraphics[width=1\textwidth]{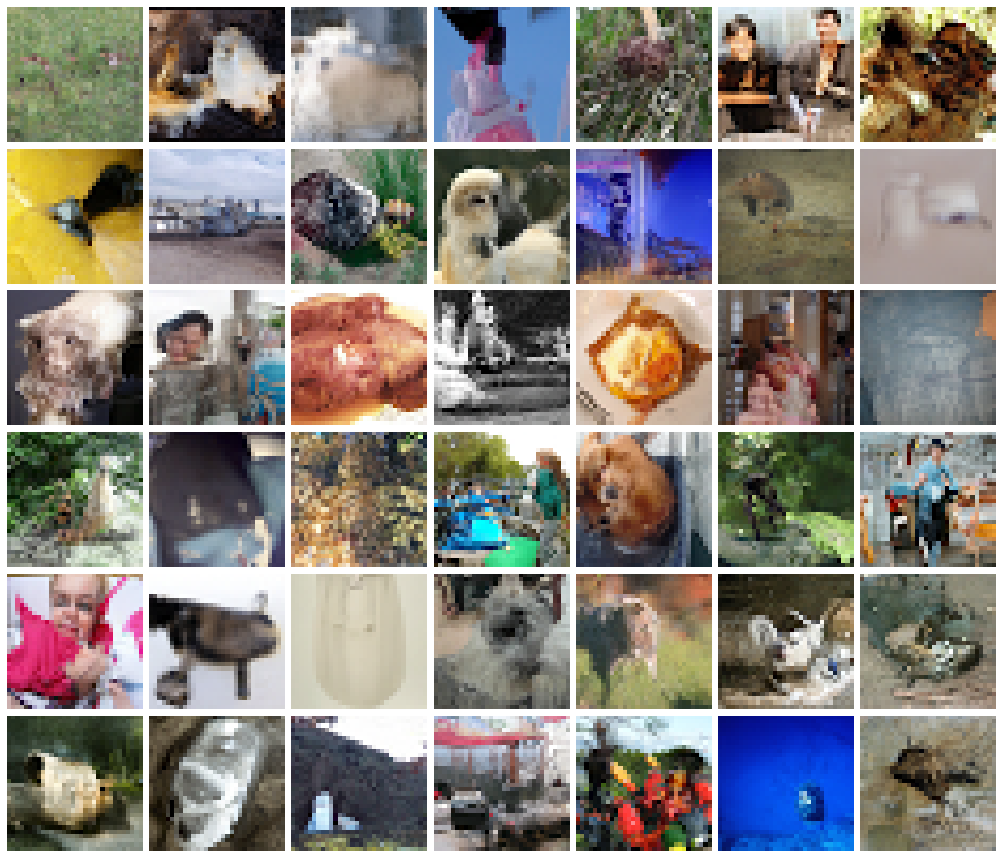}\caption{Jet (I21k) -- $32\times32$}
    \end{subfigure}
    \hspace{0.01\columnwidth}
    \begin{subfigure}[c]{0.45\columnwidth}
        \includegraphics[width=1\textwidth]{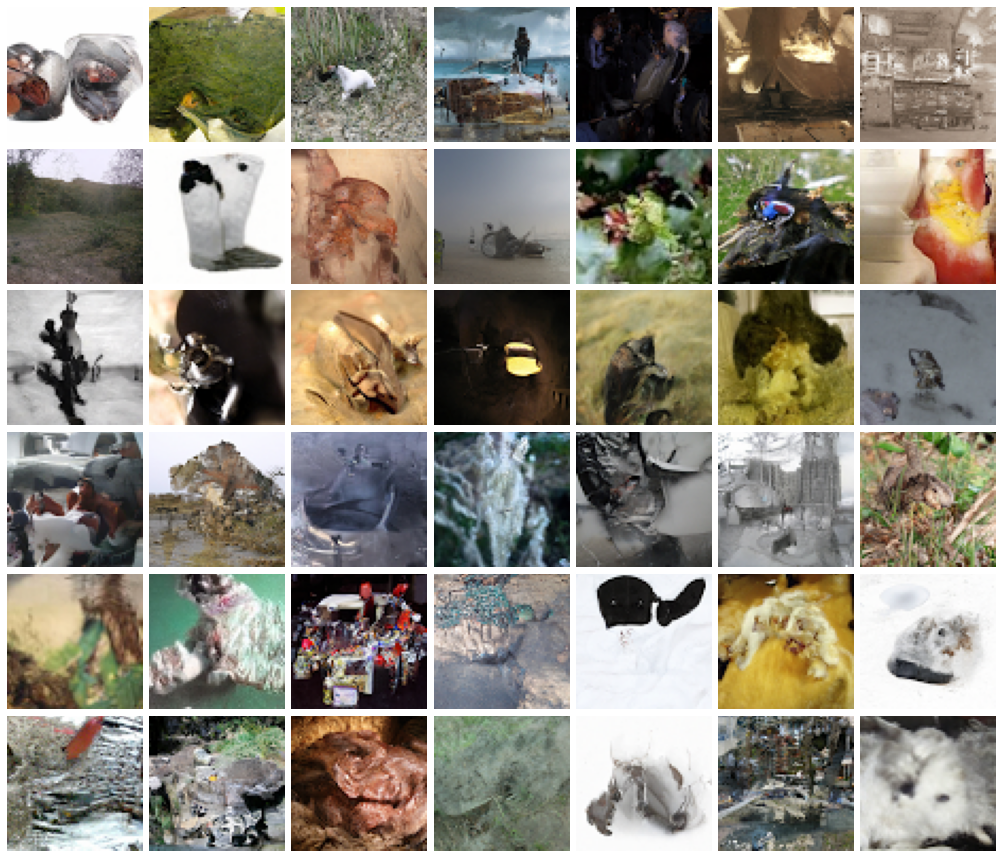}\caption{Jet (I21k) -- $64\times64$}
    \end{subfigure}
    \\
    \vspace{0.01\columnwidth}
    
    \begin{subfigure}[c]{0.45\columnwidth}
        \includegraphics[width=1\textwidth]{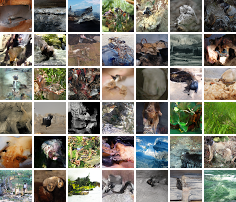}\caption{Flow++ -- $32\times32$}
    \end{subfigure}
    \hspace{0.01\columnwidth}
    \begin{subfigure}[c]{0.45\columnwidth}
        \includegraphics[width=1\textwidth]{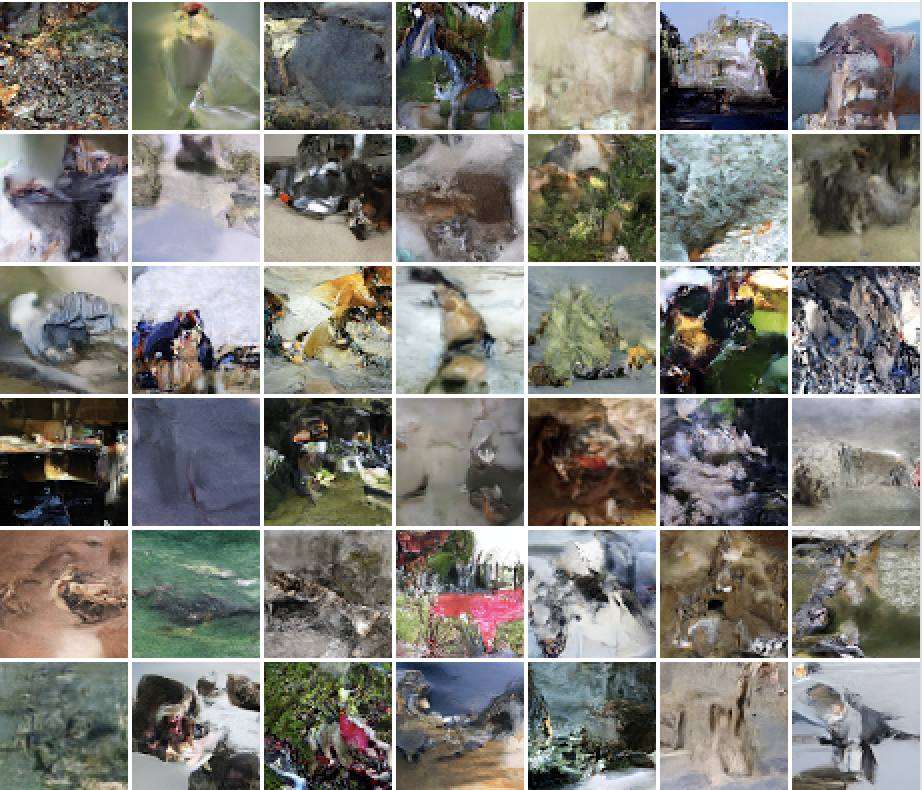}\caption{Flow++ -- $64\times64$}
    \end{subfigure}
\caption{
Random samples for ImageNet-1k at both $32\times32$ and $64\times64$ resolution. We show samples from Jet when trained from scratch and when finetuning a model pretrained on ImageNet-21k.
For comparison we also show samples from Flow++~\citep{ho2019flow++}.
}
\end{figure}

\end{document}